\documentclass[10pt,twocolumn,letterpaper]{article}

\usepackage[pagenumbers]{cvpr} 
\usepackage[acronym]{glossaries}

\usepackage{mathtools}
\usepackage{tabularx}
\usepackage{mathpazo}
\usepackage{tabularray}
\usepackage{float}
\usepackage{placeins}
\usepackage{ifthen}

\definecolor{cvprblue}{rgb}{0.21,0.49,0.74}
\usepackage[pagebackref,breaklinks,colorlinks,allcolors=cvprblue]{hyperref}

\def\paperID{13678} 
\def\confName{CVPR}
\def\confYear{2025}

\title{GreenMachine: Automatic Design of Zero-Cost Proxies for Energy-Efficient NAS}

\author{Gabriel Cortês
\and Nuno Lourenço
\and Penousal Machado
\and CISUC/LASI, DEI, University of Coimbra\\
{\tt\small \{cortes,naml,machado\}@dei.uc.pt}
}

\newacronym{ai}{AI}{Artificial Intelligence}
\newacronym[plural=ANNs,firstplural=Artificial Neural Networks (ANNs)]{ann}{ANN}{Artificial Neural Network}

\newacronym{cfg}{CFG}{Context-free Grammar}
\newacronym[plural=CPUs,firstplural=Central Processing Units (CPUs)]{cpu}{CPU}{Central Processing Unit}
\newacronym[plural=CNNs,firstplural=Convolutional Neural Networks (CNNs)]{cnn}{CNN}{Convolutional Neural Network}

\newacronym{dl}{DL}{Deep Learning}
\newacronym[plural=DNNs,firstplural=Deep Neural Networks (DNNs)]{dnn}{DNN}{Deep Neural Network}
\newacronym{denser}{DENSER}{Deep Evolutionary Network Structured Evolution}
\newacronym{dsge}{DSGE}{Dynamic Structured Grammatical Evolution}

\newacronym{ea}{EA}{Evolutionary Algorithm}
\newacronym{ec}{EC}{Evolutionary Computation}
\newacronym[plural=ES,firstplural=Evolution Strategies (ES)]{es}{ES}{Evolution Strategy}

\newacronym[plural=LLMs,firstplural=Large Language Models (LLMs)]{llm}{LLM}{Large Language Model}
\newacronym{lemonade}{LEMONADE}{Lemarckian Evolutionary Algorithm for Multi-Objective Neural Architecture Design}

\newacronym{fdenser}{Fast-DENSER}{Fast Deep Evolutionary Network Structured Representation}
\newacronym{flops}{FLOPs}{floating point operations per second}

\newacronym{ge}{GE}{Grammatical Evolution}
\newacronym{gp}{GP}{Genetic Programming}

\newacronym{iot}{IoT}{Internet of Things}

\newacronym{ml}{ML}{Machine Learning}
\newacronym{moo}{MOO}{Multi-Objective Optimization}
\newacronym{mol}{MOL}{Multi-Output Learning}
\newacronym{monas}{MONAS}{Multi-Objective Neural Architecture Search}
\newacronym{mac}{MAC}{Multiply–accumulate operation}

\newacronym{nas}{NAS}{Neural Architecture Search}
\newacronym{ne}{NE}{Neuroevolution}
\newacronym{nsga-ii}{NSGA-II}{Nondominated Sorting Genetic Algorithm II}
\newacronym{nvml}{NVML}{NVIDIA Management Library}

\newacronym[plural=SNNs,firstplural=Spiking Neural Networks (SNNs)]{snn}{SNN}{Spiking Neural Network}

\newacronym{psu}{PSU}{Power Supply Unit}

\newacronym{rnn}{RNN}{Recurrent Neural Network}

\newacronym{sge}{SGE}{Structured Grammatical Evolution}

\begin{document}

\maketitle
\begin{abstract}
\gls{ai} has driven innovations and created new opportunities across various sectors. However, leveraging domain-specific knowledge often requires automated tools to design and configure models effectively. In the case of \glspl{dnn}, researchers and practitioners usually resort to \gls{nas} approaches, which are resource- and time-intensive, requiring the training and evaluation of numerous candidate architectures. This raises sustainability concerns, particularly due to the high energy demands involved, creating a paradox: the pursuit of the most effective model can undermine sustainability goals.
To mitigate this issue, zero-cost proxies have emerged as a promising alternative. These proxies estimate a model’s performance without the need for full training, offering a more efficient approach. This paper addresses the challenges of model evaluation by automatically designing zero-cost proxies to assess \glspl{dnn} efficiently. Our method begins with a randomly generated set of zero-cost proxies, which are evolved and tested using the NATS-Bench benchmark. We assess the proxies' effectiveness using both randomly sampled and stratified subsets of the search space, ensuring they can differentiate between low- and high-performing networks and enhance generalizability. Results show our method outperforms existing approaches on the stratified sampling strategy, achieving strong correlations with ground truth performance, including a Kendall correlation of 0.89 on CIFAR-10 and 0.77 on CIFAR-100 with NATS-Bench-SSS and a Kendall correlation of 0.78 on CIFAR-10 and 0.71 on CIFAR-100 with NATS-Bench-TSS.
\end{abstract}    
\section{Introduction}
\label{sec:intro}
The impact that \gls{ai} has had on human lives and society spans various domains, from advances in healthcare diagnosis \cite{Alowais2023} to optimization of trade routes \cite{Vaddy_2023}, identification of diseases in plants \cite{Alatawi2022PlantDD}, and cyber intrusion detection \cite{Dash2022ThreatsAO}. This technology allows us to improve most fields of study. However, these advancements have a cost. Several issues affect \gls{ai} systems: racial and gender bias in automated job and loan applications \cite{Kadiresan2022}, misclassification in critical scenarios due to data manipulation by bad actors \cite{RamirezAttacks}, job displacement \cite{UnemploymentAI,GomesIRobotAIEthics}, and massive energy consumption \cite{SustainableAIWynsberghe}. The latter is quantifiable and has grown exponentially in recent years. For instance, the popular GPT-3 model, similar to the models behind OpenAI's ChatGPT, used 1287 MWh in 15 days just for its training \cite{patterson2022carbon}. The current ``arms race'' to develop \glspl{llm} models with more capabilities will likely increase the number of parameters, thereby ensuring that more recent models will consume even more energy. Moreover, serving millions of users on a daily basis requires a tremendous quantity of processing devices, such as GPUs and/or TPUs, which use a massive amount of energy. The total energy consumption of NVIDIA GPUs, the leading manufacturer of these devices, is expected to surpass the total energy needs of countries such as Belgium or Switzerland \cite{DevriesEnergy,XuLLMSurvey}. Furthermore, technological companies like Google and Microsoft, which have previously committed to offsetting their carbon emissions, have struggled to maintain these promises due to allocating more resources to \gls{ai} data centers \cite{MSNMicrosoftCarbonEmissions,Google2024EnvReport}. This growth is expected to continue, with hardly any restrictions, due to the positive impact this technology has on the population and the economy \cite{AIEconomy}.

Several techniques have been proposed to address the energy consumption problem of \gls{ai}, particularly of \gls{ml}. Reducing the floating-point precision of a model's weights has been shown to enhance energy efficiency, albeit at the tradeoff of some performance \cite{RokhQuantization}. \glspl{snn} show promising results due to their sparsity and event-based operation, though they require further hardware-side developments to fulfill their theoretical energy efficiency potential \cite{SpikingNeurons,HaiderNeuromorphic}. Using \gls{nas} or \gls{ne} to search for energy-efficient \gls{dnn} models has also significantly increased energy efficiency \cite{CortesEvoAPPS24}. This approach, however, requires substantial resources since each \gls{dnn} must undergo a full training process and subsequent evaluation, which may ultimately undermine its potential for minimal energy usage.

In this paper, we present an approach to mitigate the energy consumption problem of \gls{nas}. Zero-cost proxies estimate a \gls{dnn}'s performance without training, with some recent proxies having achieved high correlations with the ground truth, reporting a Kendall correlation of 0.706 on the NATS-Bench-TSS search space with the CIFAR-10 dataset \cite{DongParzc}, or even 0.741 on the same problem when using an ensemble of four zero-cost proxies \cite{LeeAZNAS}. This means that the score given by the zero-cost proxy to a \gls{dnn} aligns well with its actual performance in terms of accuracy.

Our main contribution is the proposal of an algorithm that automatically generates and optimizes zero-cost proxies using evolution\footnote{The code is publicly available on \href{https://github.com/RodriguesGabriel/greenmachine}{Github}}. Specifically, we begin with a randomly generated set of proxies that are then modified or evolved over several generations to gradually improve their estimation of model performance. These proxies are evaluated on the NATS-Bench benchmark across the CIFAR-10, CIFAR-100, and ImageNet16-120 datasets. We assess their performance in two ways: 1) on a subset that is a randomly selected sample of the overall search space; and 2) on a stratified subset with a diverse distribution of networks to evaluate the proxies not only on ``good'' networks but also on those with lower performance. This approach is intended to ensure that the proxies can differentiate between low- and high-performing \glspl{dnn}, minimizing overfitting and enhancing generalizability.

Experimental results indicate that this approach produced better-performing solutions than similar methods in the literature. We achieved higher correlations on nearly all datasets using a stratification technique to sample networks. Specifically, we obtained a Kendall correlation of 0.89 on CIFAR-10 and 0.77 on CIFAR-100 with NATS-Bench-SSS and a Kendall correlation of 0.78 on CIFAR-10 and 0.71 on CIFAR-100 with NATS-Bench-TSS.

This paper is structured as follows. \Cref{sec:background} presents the necessary background on \glspl{ea}, focusing on \gls{gp}, and on zero-cost proxies as well as related work. \Cref{sec:methodology} showcases the functioning of the developed algorithm and details the experimental setup. Afterward, \Cref{sec:results} present the obtained results and a discussion about them. At last, \Cref{sec:conclusion} concludes this paper and outlines potential directions for future work.
\section{Background and Related Work}
\label{sec:background}

\subsection{Genetic Programming}
\begin{figure}[h]
    \centering
    \includegraphics[width=0.9\linewidth]{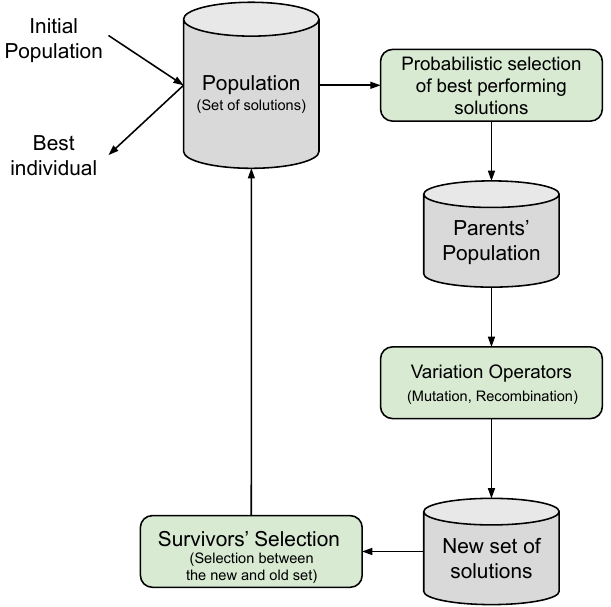}
    \caption{Simplified visualization of the flow of an \gls{ea}.}
   \label{fig:eadiagram}
\end{figure}

\gls{gp} is an \gls{ai} technique that allows for the automatic design of programs, expressions, or models with variable length, using the principles of natural selection and evolution (\cref{fig:eadiagram}) \cite{Eiben2015,WilliwsGPIntroSurvey}. When using \gls{gp}, one needs to specify the set of variables or constants and the set of functions. One of the most commonly used representations of solutions is syntax trees.

The initial set of solutions, called population, is usually created randomly. Some trees are generated with the maximum depth, others have branches with varying depths, and a mix of both approaches is also used.

Individuals are evaluated based on a fitness function that measures their performance on a given problem. The fittest individuals are then probabilistically selected to produce new solutions, or offspring, for the next generation, allowing successful traits to be passed along. Offspring undergo mutation to introduce genetic diversity, helping the population explore a broader range of potential solutions.

In the context of \gls{gp}, mutation is performed by replacing a sub-tree rooted at a randomly selected node with a newly generated sub-tree, introducing novel structures. Recombination occurs by swapping sub-trees rooted at randomly selected nodes between two individuals, enabling the exchange of genetic material and potentially beneficial traits.

\subsubsection*{Grammatical Evolution}
\gls{ge} is a grammar-based \gls{gp} that uses a variable-length integer representation  \cite{RyanGE}. It relies on the production rules of a \gls{cfg} to generate solutions. A \gls{cfg} is defined by a tuple $( N, T, P, S )$ with $N$ being the set of non-terminals, $T$ the set of terminals, $P$ the set of production rules, and $S \in N$ is a start symbol.

The solution generation process from an integer sequence is carried out as follows: the sequence is read from left to right, beginning with $S$, and a production rule is iteratively applied to expand the leftmost non-terminal symbol. The current item's value in the sequence determines the production rule to use by calculating its modulo with respect to the number of available expansions for the leftmost non-terminal symbol.

Consider the grammar in \cref{grammar} with $N = \{\texttt{<expr>}, \texttt{<op>}, \texttt{<trig>}, \texttt{<var>}\}$, $T = \{+, -, \sin, \cos, x, y, (, )\}$ and S = \texttt{<expr>}. Considering a genotype with integer values ranging from 0 to 63, the mapping process is as follows (\cref{GEMapping}). First, we begin with the start symbol, \texttt{<expr>}, and read the first element of the integer vector. Since there are three options on how to expand the start symbol, we have $46 \mod 3 = 1$; thus, the start symbol is replaced by its second expansion rule, \texttt{<trig>(<expr>)}. Following this, the second value in the genotype is read, and we expand the leftmost non-terminal symbol, repeating the process. This is executed until there are no more non-terminal symbols. It is possible to use a wrapping mechanism instead of terminating the mapping process if there are no more integers to read from the genotype, thus reusing the genotype values.

\begin{figure}[ht]
	\centering
	{\setlength\tabcolsep{4pt}\centering
		\begin{tabular}{>{$}l<{$}>{$}r<{$}>{$}l<{$}}
			$\texttt{<expr>}$ & \Coloneqq & $\texttt{<expr>} \texttt{<op>} \texttt{<expr>}$ \\
			         & |         & $\texttt{<trig>(<expr>)}$     \\
			         & |         & $\texttt{<var>}$
			\\
			$\texttt{<op>}$   & \Coloneqq & $+$ \hspace{5pt} | \hspace{5pt} $-$                \\
			$\texttt{<trig>}$ & \Coloneqq & \texttt{sin} \hspace{5pt} | \hspace{5pt} \texttt{cos}                \\
			$\texttt{<var>}$  & \Coloneqq & $\texttt{x}$ \hspace{5pt} | \hspace{5pt} $\texttt{y}$                 \\
		\end{tabular}}
	\caption{Example of a grammar.}
	\label{grammar}
\end{figure}

\begin{figure}[ht]
	\centering
	Genotype: [46, 15, 17, 28, 50, 42, 22, 19, 51, 35] \\[0.3cm]

	\begin{tabular}{cr}
		\texttt{<expr>}         &                 \\[0.1cm]
		\texttt{<trig>(<expr>)} & $46 \mod 3 = 1$ \\[0.1cm]
		$\texttt{cos(<expr>)}$ & $15 \mod 2 = 1$ \\[0.1cm]
		$\texttt{cos(<var>)}$  & $17 \mod 3 = 2$ \\[0.1cm]
		$\texttt{cos(x)}$      & $28 \mod 2 = 0$
	\end{tabular}
	\caption[Example of GE mapping.]{Example of Grammatical Evolution mapping.}
	\label{GEMapping}
\end{figure}

Despite \gls{ge}'s flexibility and ease of use, it has suffered from issues, such as low locality and high redundancy \cite{LourencoSGE}. To address these limitations, \gls{sge} \cite{LourencoDSGE} was introduced. \gls{sge} mitigates these issues by associating each gene with a specific non-terminal and using variable-length integer lists to represent expansion choices. This structure ensures that changing one gene does not affect the derivation choices for other non-terminals, which reduces phenotypic changes and thereby improves locality. Furthermore, the values in each list are constrained by the number of expansion options available for the corresponding non-terminal, eliminating the need for modulo operations and reducing redundancy. Empirical results demonstrate that \gls{sge} outperforms \gls{ge} across various optimization problems.

\subsection{Zero-cost proxies}
Zero-Shot, also known as the training-free method, is a technique that allows the prediction of the quality of a model without training it \cite{TrainingFreeNASReview}. This is achieved through proxies, which are algorithms or mathematical formulas that estimate how good a model might be. These proxies allow us to assess the performance of a \gls{dnn} model without training it, thus saving resources. Zero-cost proxies are a relatively recent research thread in the \gls{ml} community since they were introduced in 2018 by Camero et al. \cite{CameroLowCostExpectedPerformance} and have since been continuously improved. Traditional \gls{nas} methods typically require hundreds or thousands of GPU hours and, as such, can significantly benefit from using training-free methods. Zero-cost proxies allow us to predict the performance of a \gls{dnn} model without training it, and they usually require only a small amount of GPU time or even CPU time to do so. Despite numerous statistical analyses of training-free NAS algorithms, a theoretical analysis of training-free algorithms remains lacking. A comprehensive theoretical examination of this score function is essential to further research in this field \cite{TrainingFreeNASReview}.

To avoid training many networks to evaluate the correlation between the proxy and the actual accuracy, well-established \gls{nas} datasets are used \cite{TrainingFreeNASReview}. Among other metrics, these datasets contain the architectures of many networks and their corresponding accuracies.

A pioneer in the field, NASWOT is an algorithm that generates scores reflecting a model’s test accuracy without requiring training. It computes these scores based on the network’s activation patterns in response to a single mini-batch \cite{MellorNASWOT}. Synflow is a pruning algorithm that aims to prevent the layer collapse problem when pruning a neural network. It was extended as a data-independent estimator of a network's performance without training it \cite{Synflow}. Inspired by pruning-at-initialization, GradNorm is a proxy metric that computes the sum of the Euclidean norms of the gradients after passing a single mini-batch of training data \cite{Gradnorm}. TE-NAS is a framework that evaluates network architectures by examining the spectrum of the neural tangent kernel and the number of linear regions in the input space \cite{TENAS}. Zen-NAS is a zero-shot method that measures the expressivity of a \gls{dnn} by computing its Zen-Score, which is derived from a few forward inferences on randomly initialized networks with random Gaussian inputs \cite{ZenNAS}. ZiCo demonstrates that high-performance \glspl{dnn} tend to possess high absolute mean values and low standard deviation values for the gradient and uses that information to estimate the performance of a network \cite{Zico}. 

EZNAS proposes a \gls{gp} approach to automate the discovery of zero-cost proxies for \gls{nas} scoring \cite{EZNAS} by using the DEAP framework \cite{DEAP}. Each individual's fitness is determined by the minimum Kendall $\tau$ coefficient it achieves across both search spaces. After the evolutionary process, the fittest individual is evaluated on 4000 random networks from the same two search spaces. While EZNAS's results are competitive with the current state of the art, the authors do not provide details on how the other zero-cost proxies were evaluated, making direct comparisons potentially unfair. Furthermore, EZNAS uses statistics from only a portion of the overall layers of the models and employs a relatively low recombination rate of 40\%, which may not be sufficient to generate optimal individuals consistently.
\section{Methodology}
\label{sec:methodology}

\subsection{Benchmarks}
Benchmarking \gls{nas} algorithms is challenging due to variations in data preprocessing, evaluation pipelines, and even random seeds. Moreover, fully training a large number of networks requires extensive computational resources, thus making reproducibility difficult for most researchers. To address this issue, some datasets were proposed to standardize the benchmarking process, providing a common ground for comparisons and reducing the required computational resources by delivering the results that would otherwise require the complete training of many \gls{dnn} models.

These benchmarks typically include not only the network configuration and its accuracy metric after training but also other data such as latency, number of parameters, and more, enabling a quicker assessment of the proposed \gls{nas} method.

NAS-Bench-201 features full graph cells, allowing for a more comprehensive search space, though limited to four nodes and five associated operation options to maintain manageability \cite{NASBench201}. Each architecture is trained and evaluated on the CIFAR-10, CIFAR-100, and ImageNet-16-120 image classification datasets. It contains 15,625 architectures. NATS-Bench is a unified benchmark for searching for architecture topology and size \cite{NATSBench}. It includes 15,625 candidates for the architecture topology (TSS) and 32,768 for architecture size (SSS). It also presents results for CIFAR-10, CIFAR-100, and ImageNet-16-120. Although not explicitly mentioned in the paper, the authors note in the project's Github repository\footnote{\href{https://github.com/D-X-Y/NATS-Bench}{https://github.com/D-X-Y/NATS-Bench}} that the topology search space is equivalent to NAS-Bench-201.

\subsection{Experimental Setup}
The experiments were conducted on a machine running Ubuntu 22.04.3 LTS with two Intel Xeon Silver 4310 CPUs with a clock frequency of 2.10GHz and 12 cores each, 256 GB of RAM, and three NVIDIA RTX A6000 GPUs with 48 GB of GDDR6 RAM. The environment in which they were executed used CUDA 12.1, Python 3.10, and PyTorch 2.3.1.

\subsubsection{Networks Sampling}
\begin{figure}[ht]
    \centering
    \includegraphics[width=\linewidth]{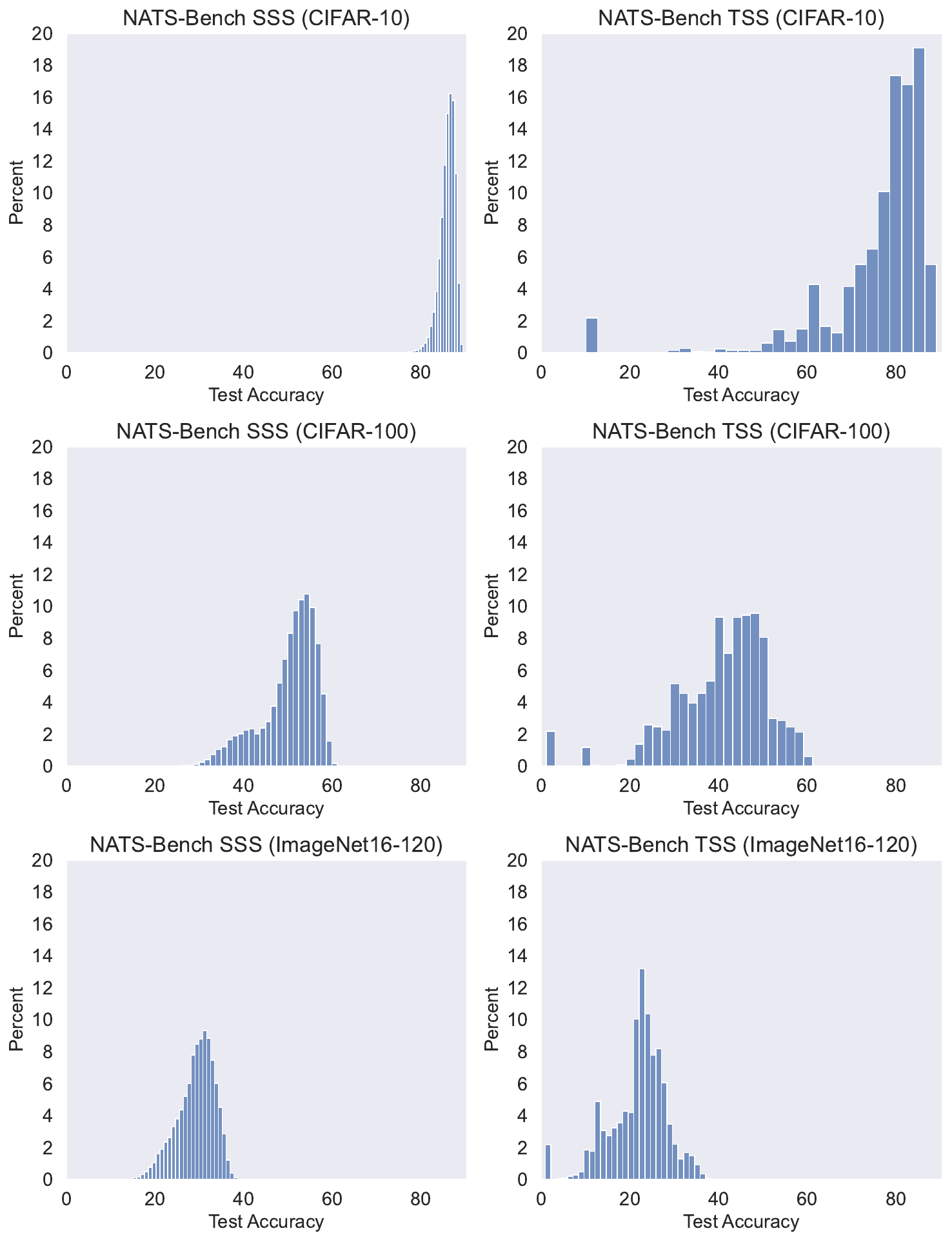}
    \caption{Histograms of test accuracy for the NATS-Bench benchmark across the topology and size search spaces and the CIFAR-10, CIFAR-100, and ImageNet16-120 datasets.}
   \label{fig:nats_bench_histogram}
\end{figure}

After analyzing the search spaces (\cref{fig:nats_bench_histogram}), we observed a skew towards networks with high accuracy across both search spaces, regardless of the dataset. Consequently, random sampling does not ensure adequate search space coverage, as it may produce a sample set that is too constrained or biased toward high-performance networks. This bias can result in proxies that fail to distinguish effectively between low- and high-performing networks. To address this issue, we stratified the search space based on test accuracy, creating five groups or bins of networks according to their performance. This approach ensures a diverse set of samples for evaluation. Additionally, we sample 20 networks from each dataset for each search space using this stratification method to maintain a balanced representation across different performance bins.

\subsubsection{Zero-Cost Proxies Evaluation}
We assess the usefulness of the zero-cost proxies using search spaces that comprise architectures of \glspl{dnn} and their respective test accuracy after training. Specifically, we use the NATS-Bench benchmark \cite{NATSBench}. The evaluation uses the zero-cost proxy to score every network sampled. Then we calculate Kendall's correlation ($\tau$) between the set of scores ($S$) and the actual test accuracy ($A$). The $\tau$ measures the ordinal association between the two sets according to the following:

Let \( S = \{s_1, s_2, \dots, s_n\} \) represent the scores from a zero-cost proxy, and \( A = \{a_1, a_2, \dots, a_n\} \) represent the test accuracies.

The Kendall tau correlation coefficient between \( S \) and \( A \) is given by:

\[
\tau = \frac{P - Q}{\sqrt{(P + Q + T)(P + Q + U)}}
\]

where:

\begin{itemize}
    \item \( P \) is the number of concordant pairs, i.e., pairs \((s_i, a_i)\) and \((s_j, a_j)\) such that \((s_i - s_j)(a_i - a_j) > 0\),
    \item \( Q \) is the number of discordant pairs, i.e., pairs \((s_i, a_i)\) and \((s_j, a_j)\) such that \((s_i - s_j)(a_i - a_j) < 0\),
    \item \( T \) is the number of ties in \( S \), i.e., pairs \((s_i, s_j)\) such that \( s_i = s_j \),
    \item \( U \) is the number of ties in \( A \), i.e., pairs \((a_i, a_j)\) such that \( a_i = a_j \).
\end{itemize}

This metric indicates the zero-cost proxy’s suitability for predicting model performance, reflecting the likelihood that a proxy score correlates with the actual model accuracy. 

The evaluation function is defined as the sum of the absolute values of the Kendall rank correlation coefficients across all search space and dataset combinations. We use absolute values rather than signed ones to detect any correlation, not solely a positive one.

\subsubsection{Feature Extraction}
We extract 20 features from each layer of a network. When applicable, these are the weights and gradients of the layer and the weights before and after a forward pass or a backward pass takes place. At first, we consider the randomly initialized network and archive each layer's weights and gradients. After this, we repeat the complete extraction process in three modes: one where we pass a batch of random data on the network, another where we pass a batch from the dataset, and, last, a mode where we pass a batch from the dataset but perturbed with noise.

Having this archive of network statistics, we then iterate over each of the network's layers and compute the current individual's formula. The final score attributed to the network is the mean value of the score of all layers.

\subsubsection{Operations}
The available operations range from essential mathematical functions like addition, subtraction, element-wise product, and matrix multiplication to specialized computations such as Frobenius norm and eigenvalue ratios. We also include activation functions like ReLU and sigmoid and normalization techniques. Additionally, we provide methods for noise addition, cosine similarity, and logical comparisons. The complete list of operations is presented as supplementary material. 

To handle feature extraction from every layer, we ensure that matrices of any dimension are compatible for operations. We do this by flattening the matrices, comparing their lengths, and padding the shorter one with a constant (1). This gives both matrices the same number of elements, allowing them to be reshaped back to the original dimensions of the larger matrix. Additionally, we replace invalid tensors with a default tensor of 1 and substitute invalid values within tensors to maintain valid operations and prevent runtime errors, ensuring that no individuals are excluded due to shape mismatches.

\subsubsection{Search Algorithm}
To perform the search for zero-cost proxies, we used the grammar-based \gls{gp} approach described in \cite{LourencoDSGE}, with the parameters detailed in \cref{tab:experimental_evozero:params}. Specifically, we performed five runs, each for 100 generations, with a population size of 100 individuals. To preserve the best solutions, an elitism size of 10 was applied. Selection was based on tournament selection with a tournament size of 5. The genetic operators were configured with a crossover rate of 90\% and a mutation rate of 50\%. The evolutionary trees had depths ranging from 5 to 12, and each experiment evaluated 120 networks.

\Cref{tab:experimental_evozero:params} lists the used experimental parameters.

\begin{table}
\centering
\caption{Experimental parameters.}
\label{tab:experimental_evozero:params}
\begin{tblr}{
  rowsep = 0.6pt,
  cell{2}{2} = {c},
  cell{3}{2} = {c},
  cell{4}{2} = {c},
  cell{5}{2} = {c},
  cell{6}{2} = {c},
  cell{7}{2} = {c},
  cell{8}{2} = {c},
  cell{9}{2} = {c},
  cell{10}{2} = {c},
  hline{1,11} = {-}{0.08em},
  hline{2} = {-}{},
}
Evolutionary Parameter     & Value     \\
Number of runs             & 5         \\
Number of generations      & 100       \\
Population size            & 100       \\
Elitism size               & 10        \\
Tournament size            & 5         \\
Crossover rate             & 90\%      \\
Mutation rate              & 50\%      \\
Tree depth                 & {[}5, 12] \\
Num. of evaluated networks & 120       
\end{tblr}
\end{table}
\section{Results and Discussion}
\label{sec:results}
\begin{figure}[ht]
    \centering
    \includegraphics[width=\linewidth]{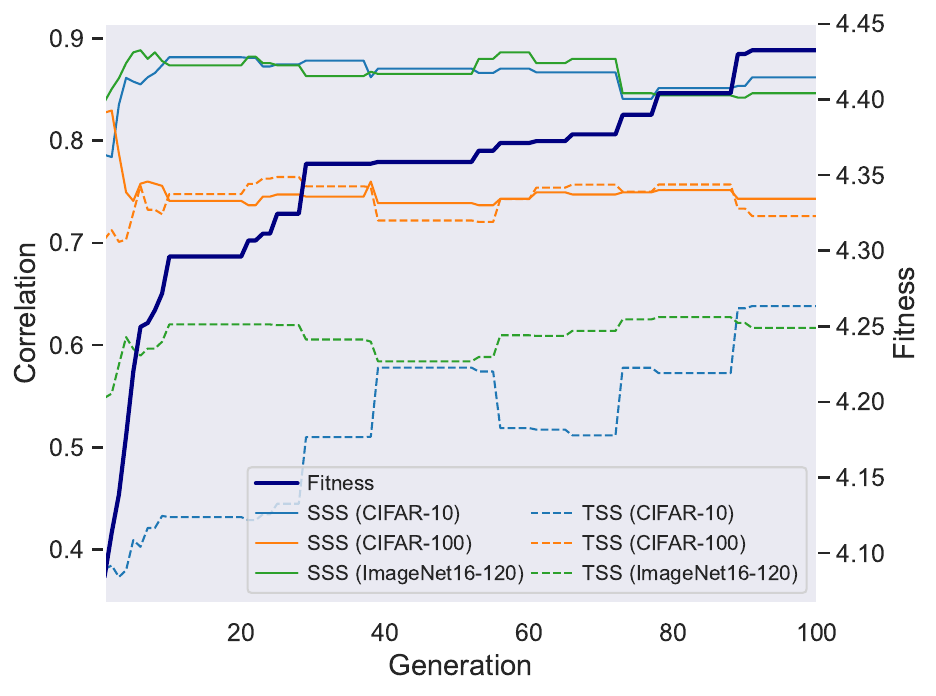}
    \caption{Evolution of the Kendall correlation coefficient on NATS-Bench's two search spaces and the fitness metric over 100 generations, averaged over 5 runs.}
   \label{fig:evolution}
\end{figure}

\Cref{fig:evolution} shows the evolution of zero-cost proxies’ performance, measured by the Kendall correlation coefficient ($\tau$), over 100 iterations across NATS-Bench’s two search spaces on the CIFAR-10, CIFAR-100, and ImageNet16-120 datasets, averaged over five runs. In the figure, dashed lines represent performance on the Topology Search Space (TSS), while solid lines indicate performance on the Size Search Space (SSS). The thick solid line illustrates the fitness of the individuals, defined as the sum of the absolute values of the $\tau$ scores across the datasets.

Examining the results in \cref{fig:evolution}, we observe that zero-cost performance on the CIFAR-10 and ImageNet16-120 datasets in the TSS (dashed lines) is lower than in the SSS (solid lines). For CIFAR-100, the difference is minimal. This suggests that identifying zero-cost proxies in the TSS is more challenging than in the SSS. This difficulty can be attributed to the typical correlation between network size and performance \cite{hestness2017deeplearningscalingpredictable}. A proxy that leverages the number of network parameters to estimate performance has a better chance of accurately predicting performance in the SSS. In contrast, extracting performance-related features based on network topology is inherently more complex, which is required to estimate the performance on the TSS.

Additionally, as shown in \cref{fig:nats_bench_histogram}, the distribution of network performance differs between the search spaces. In the TSS, network performance ranges widely from about 10\% (close to random choice) to nearly 95\%, while in the SSS, performance is concentrated within narrower accuracy ranges.

Finally, slight decreases in the solid and dashed lines can be observed, particularly in the correlations measured in the SSS search space. Since the quality of the proxies is defined as the sum of the $\tau$ correlations across both search spaces, this behavior is expected, as tradeoffs are made to improve correlation in the other search space.

\paragraph{Validating the Zero-Cost Proxies}

To assess the generalization ability of the zero-cost proxies, we evaluate them on networks that differ from those used in training. Specifically, we test the proxies on 4,500 new networks. We evaluate 30 sets of 150 networks from each search space and dataset, using stratified and non-stratified sampling strategies. This approach ensures that our evaluation covers a diverse range of architectures, allowing us better to assess the generalization and robustness of the proxies. With 30 sets of networks, we report the mean and standard deviation across these sets.

\begin{figure}
    \centering
    \includegraphics[width=\linewidth]{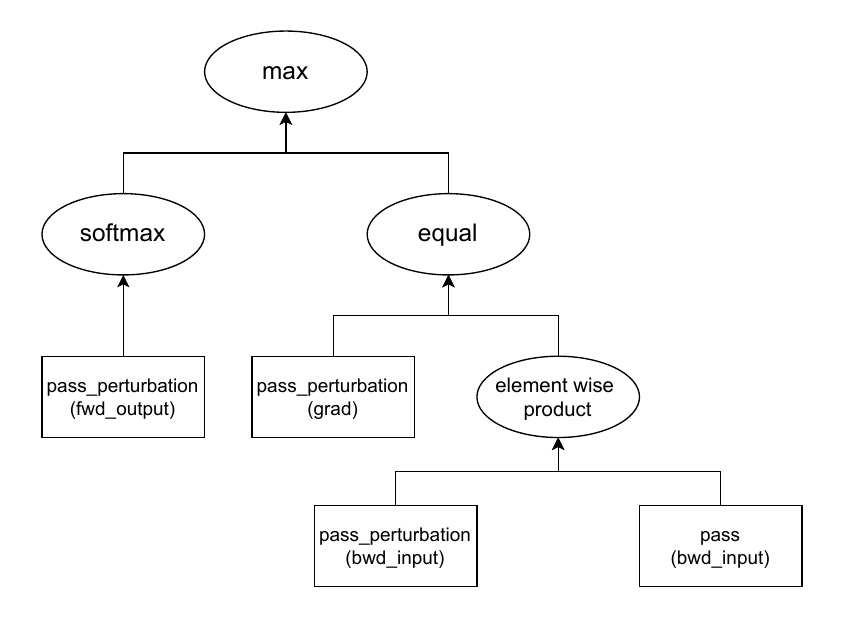}
    \caption{Representation of the GreenMachine-2 solution, where ellipses represent functions and rectangles denote terminal symbols (variables).}
    \label{fig:greenmachine_2}
\end{figure}

We implemented state-of-the-art zero-cost proxies from the literature and applied them to the same set of networks to ensure a fair comparison. Additionally, we included a random proxy that generates a random score between 0 and 1 for comparison purposes. \Cref{fig:greenmachine_2} depicts the GreenMachine-2 solution, and the formulas for the remaining best solutions discovered by our approach are provided as supplementary material.

\Cref{table:resultsnon_stratified_kendall,table:resultsstratified_kendall} present the validation results for the two sampling strategies using the $\tau$ correlation coefficient. The corresponding results for Spearman’s rank correlation are available as supplementary material.

In \cref{table:resultsnon_stratified_kendall}, where networks are randomly sampled without considering performance representativity, we observe no significant differences among the zero-cost proxies. However, one zero-cost proxy discovered by our approach, GreenMachine-3 (GM-3), outperforms the others on CIFAR-100 when applied to the SSS.

This result can be explained by the fact that the randomly sampled set of networks shows very similar performances, as indicated in \cref{table:performancestatsvalidation}. As mentioned previously, if network sampling is not done carefully, the resulting set can consist of models with very similar performance. Observing the standard deviation of the non-stratified set of networks, we see that it is low, indicating high similarity in performance. This reduces the need for proxies to make clear distinctions. However, when evaluating CIFAR-100 on the SSS, the standard deviation increases, requiring proxies to distinguish between high- and low-performing networks more effectively.

\begin{table}[ht!]
\centering
\caption{Characterization of the networks used for validation of zero-cost proxies. Each cell presents the average accuracy (\%) and standard deviation of 30 sets of 150 networks for the non-stratified and stratified versions of the SSS and TSS search spaces across the three different datasets.}
\label{table:performancestatsvalidation}
\resizebox{\linewidth}{!}{%
\begin{tblr}{
  column{even} = {c},
  column{3} = {c},
  cell{2}{2} = {c=3}{},
  cell{5}{2} = {c=3}{},
  hline{1,8} = {-}{0.08em},
  hline{2-3,5-6} = {-}{},
}
    & \textbf{CIFAR-10} & \textbf{CIFAR-100} & \textbf{ImageNet16-120} \\
    & Non-Stratified &           &                \\
SSS & $85.98 \pm 1.74$      & $50.13 \pm 6.28$ & $29.19 \pm 4.09$      \\
TSS & $75.83 \pm 13.76$      & $40.60 \pm 11.06$ & $21.74 \pm 6.36$      \\
    & Stratified     &           &                \\
SSS & $80.71 \pm 4.90$      & $42.95 \pm 9.71$ & $26.47 \pm 6.59$      \\
TSS & $49.28 \pm 25.15$      & $31.15 \pm 17.21$ & $19.04 \pm 10.96$      
\end{tblr}
}
\end{table}

Concerning the validation where we used a stratified sampling strategy, the zero-cost proxies discovered by our approach can clearly distinguish between low-performing and high-performing networks (see \cref{table:resultsnon_stratified_kendall}). Our proxies achieve the highest $\tau$ correlation across the entire SSS search space, and, on the TSS search space, our solutions surpass those in the literature in all cases except for the ImageNet16-120 dataset. This exception can be explained by the distribution of network performances: as shown in \cref{table:performancestatsvalidation}, the TSS ImageNet16-120 stratified version has a low standard deviation, indicating that network performances are very similar in this sample set, and proxies are not required to distinguish between low and high performers.

\begin{table*}[ht]
\centering
\caption{Comparison of Zero-Cost proxies on the NATS-Bench benchmark across the CIFAR-10, CIFAR-100, and ImageNet16-120 datasets on the \textbf{non-stratified subset}. The presented values are the \textbf{mean absolute Kendall correlation coefficient} over 30 runs, multiplied by 100. Bold denotes the best value.}
\label{table:resultsnon_stratified_kendall}
\resizebox{\linewidth}{!}{%
\begin{tblr}{
  colsep=4pt,
  rows={0.5mm},
  hline{1,2,14} = {-}{0.08em},
  hline{11} = {-}{},
  colspec = {Q[l] Q[c] Q[c] Q[c] Q[c] Q[c] Q[c]},
}
Proxy & SSS (CF-10) & SSS (CF-100) & SSS (IN16-120) & TSS (CF-10) & TSS (CF-100) & TSS (IN16-120)  &        \\
 Random	& 4.0 $\pm$ 2.2	& 4.4 $\pm$ 3.9	& 4.5 $\pm$ 3.6	& 4.4 $\pm$ 3.4	& 4.2 $\pm$ 2.8	& 4.1 $\pm$ 3.1\\
\#Params	& 67.4 $\pm$ 3.3	& 53.5 $\pm$ 4.5	& 65.6 $\pm$ 3.3	& 37.3 $\pm$ 4.6	& 35.1 $\pm$ 5.5	& 28.3 $\pm$ 5.1\\
Synflow \cite{Synflow}	& \textbf{76.8 $\pm$ 1.9}	& 59.0 $\pm$ 4.1	& \textbf{79.5 $\pm$ 2.3}	& 37.4 $\pm$ 5.7	& 35.1 $\pm$ 5.7	& 35.6 $\pm$ 4.6\\
Gradnorm \cite{Gradnorm}	& 18.7 $\pm$ 5.0	& 48.4 $\pm$ 4.8	& 36.8 $\pm$ 5.8	& 14.7 $\pm$ 5.9	& 5.9 $\pm$ 4.4	& 6.7 $\pm$ 3.8\\
NASWOT \cite{MellorNASWOT}	& 38.1 $\pm$ 4.2	& 16.1 $\pm$ 5.3	& 40.7 $\pm$ 4.2	& 43.0 $\pm$ 4.6	& 38.0 $\pm$ 4.3	& 38.5 $\pm$ 5.8\\
TE-NAS \cite{TENAS}	& 34.2 $\pm$ 4.0	& 28.1 $\pm$ 5.8	& 38.8 $\pm$ 5.3	& 24.8 $\pm$ 4.8	& 13.0 $\pm$ 5.7	& 8.2 $\pm$ 6.0\\
Zen-NAS \cite{ZenNAS}	& 72.6 $\pm$ 2.5	& 47.0 $\pm$ 5.0	& 67.0 $\pm$ 2.9	& 12.6 $\pm$ 5.1	& 17.0 $\pm$ 5.8	& 21.9 $\pm$ 4.8\\
ZiCo \cite{Zico}	& 73.1 $\pm$ 2.1	& 56.3 $\pm$ 3.1	& 73.5 $\pm$ 2.8	& 39.0 $\pm$ 5.5	& 35.6 $\pm$ 5.3	& 36.0 $\pm$ 4.8\\
EZNAS \cite{EZNAS}	& 72.5 $\pm$ 2.5	& 48.7 $\pm$ 4.1	& 57.0 $\pm$ 2.9	& \textbf{61.1 $\pm$ 3.5}	& \textbf{60.9 $\pm$ 3.8}	& \textbf{54.7 $\pm$ 4.6}\\
GM-1 (Ours)	& 51.8 $\pm$ 3.2	& 55.9 $\pm$ 4.3	& 49.7 $\pm$ 3.8	& 50.2 $\pm$ 4.8	& 48.1 $\pm$ 4.5	& 39.1 $\pm$ 4.8\\
GM-2 (Ours)	& 68.5 $\pm$ 2.9	& 62.4 $\pm$ 3.9	& 78.0 $\pm$ 1.7	& 33.7 $\pm$ 5.2	& 31.0 $\pm$ 4.9	& 31.5 $\pm$ 4.7\\
GM-3 (Ours)	& 75.6 $\pm$ 2.5	& \textbf{66.8 $\pm$ 3.3}	& 70.2 $\pm$ 2.7	& 8.9 $\pm$ 4.9	& 6.7 $\pm$ 4.4	& 28.8 $\pm$ 5.0
\end{tblr}
}
\end{table*}

\begin{table*}[ht]
\centering
\caption{Comparison of Zero-Cost proxies on the NATS-Bench benchmark across the CIFAR-10, CIFAR-100, and ImageNet16-120 datasets on the \textbf{stratified subset}. The presented values are the \textbf{mean absolute Kendall correlation coefficient} over 30 runs, multiplied by 100. Bold denotes the best value.}
\label{table:resultsstratified_kendall}
\resizebox{\linewidth}{!}{%
\begin{tblr}{
  colsep=4pt,
  rows={0.5mm},
  hline{1,2,14} = {-}{0.08em},
  hline{11} = {-}{},
  colspec = {Q[l] Q[c] Q[c] Q[c] Q[c] Q[c] Q[c]},
}
Proxy & SSS (CF-10) & SSS (CF-100) & SSS (IN16-120) & TSS (CF-10) & TSS (CF-100) & TSS (IN16-120)  &        \\
 Random	& 5.1 $\pm$ 3.4	& 4.5 $\pm$ 2.9	& 4.2 $\pm$ 3.7	& 4.3 $\pm$ 3.3	& 3.4 $\pm$ 2.9	& 4.0 $\pm$ 3.0\\
\#Params	& 75.8 $\pm$ 2.4	& 48.9 $\pm$ 4.5	& 59.4 $\pm$ 3.5	& 38.1 $\pm$ 3.7	& 42.9 $\pm$ 4.8	& 31.3 $\pm$ 4.1\\
Synflow \cite{Synflow}	& 80.6 $\pm$ 1.8	& 59.8 $\pm$ 3.9	& 71.3 $\pm$ 2.8	& 73.9 $\pm$ 2.1	& 45.9 $\pm$ 3.1	& 39.9 $\pm$ 4.7\\
Gradnorm \cite{Gradnorm}	& 41.0 $\pm$ 3.9	& 59.1 $\pm$ 2.2	& 47.4 $\pm$ 3.6	& 59.5 $\pm$ 2.4	& 20.9 $\pm$ 5.0	& 7.1 $\pm$ 4.9\\
NASWOT \cite{MellorNASWOT}	& 53.3 $\pm$ 3.5	& 14.9 $\pm$ 7.1	& 34.7 $\pm$ 5.1	& 71.2 $\pm$ 2.5	& 52.7 $\pm$ 3.2	& 43.2 $\pm$ 4.0\\
TE-NAS \cite{TENAS}	& 53.8 $\pm$ 2.8	& 28.6 $\pm$ 6.2	& 36.3 $\pm$ 4.6	& 14.5 $\pm$ 5.3	& 21.0 $\pm$ 7.2	& 11.6 $\pm$ 5.1\\
Zen-NAS \cite{ZenNAS}	& 77.2 $\pm$ 1.9	& 47.2 $\pm$ 4.7	& 59.0 $\pm$ 3.7	& 31.0 $\pm$ 4.9	& 18.2 $\pm$ 5.7	& 22.0 $\pm$ 5.7\\
ZiCo \cite{Zico}	& 79.5 $\pm$ 1.9	& 58.1 $\pm$ 3.7	& 80.9 $\pm$ 1.2	& 75.0 $\pm$ 2.2	& 48.3 $\pm$ 3.5	& 62.1 $\pm$ 2.6\\
EZNAS \cite{EZNAS}	& 82.8 $\pm$ 0.9	& 64.4 $\pm$ 2.6	& 70.7 $\pm$ 2.1	& 57.4 $\pm$ 2.3	& 70.0 $\pm$ 2.2	& \textbf{66.2 $\pm$ 3.4}\\
GM-1 (Ours)	& 61.9 $\pm$ 2.6	& 65.3 $\pm$ 2.9	& 67.6 $\pm$ 2.1	& \textbf{78.3 $\pm$ 2.0}	& \textbf{70.5 $\pm$ 2.2}	& 64.3 $\pm$ 3.0\\
GM-2 (Ours)	& 82.7 $\pm$ 1.4	& \textbf{76.9 $\pm$ 1.6}	& \textbf{85.6 $\pm$ 0.9}	& 65.5 $\pm$ 2.9	& 55.8 $\pm$ 3.2	& 55.8 $\pm$ 3.5\\
GM-3 (Ours)	& \textbf{88.8 $\pm$ 0.9}	& 74.4 $\pm$ 1.3	& 79.4 $\pm$ 1.3	& 39.1 $\pm$ 4.5	& 29.6 $\pm$ 4.6	& 57.0 $\pm$ 3.4
\end{tblr}
}
\end{table*}

\section{Conclusion}
\label{sec:conclusion}
This paper introduces GreenMachine, an algorithm that automatically designs zero-cost proxies using an evolutionary approach. We evaluate the effectiveness of the discovered proxies using the NATS-Bench benchmark across CIFAR-10, CIFAR-100, and ImageNet16-120 datasets. 

The solutions discovered by our proposed approach perform better than existing zero-cost proxy methods when distinguishing between low- and high-performing networks. To assess this, we apply stratified sampling to the search space, dividing it into groups based on test accuracy to ensure a representative set of samples for evaluating the proxies. We use the Kendall correlation coefficient to measure the correlation between the proxy scores and network accuracy.

The results show that we achieve better performance on nearly all datasets, demonstrating the ability to effectively differentiate between low- and high-performing \glspl{dnn} while minimizing overfitting and enhancing generalizability. Notably, on the CIFAR-10 dataset using the NATS-Bench-SSS benchmark, GreenMachine achieved a Kendall correlation of 0.89, and on the NATS-Bench-TSS benchmark with the same dataset, a Kendall correlation of 0.78.

\subsection{Future Work}
In this work, we evolved and tested the generated solutions on only two benchmarks. Using other search spaces and datasets could enhance the generalizability of our results, allowing for a better assessment of the proxy's effectiveness across multiple problem domains.

Allied with evaluating the individuals on more search spaces, it might be relevant to experiment using more specialized selection operators, such as lexicase selection \cite{SpectorLexicase}, to promote solutions that perform well across a diverse range of tasks.

One potential enhancement to the evolutionary algorithm involves using ephemeral constants, which could fine-tune the relative importance of features within the solutions, improving the quality of the proxies. 
{
    \small
    \bibliographystyle{ieeenat_fullname}
    \bibliography{main}

\begin{thebibliography}{40}
\providecommand{\natexlab}[1]{#1}
\providecommand{\url}[1]{\texttt{#1}}
\expandafter\ifx\csname urlstyle\endcsname\relax
  \providecommand{\doi}[1]{doi: #1}\else
  \providecommand{\doi}{doi: \begingroup \urlstyle{rm}\Url}\fi

\bibitem[Abdelfattah et~al.(2021)Abdelfattah, Mehrotra, Dudziak, and
  Lane]{Gradnorm}
Mohamed~S. Abdelfattah, Abhinav Mehrotra, Lukasz Dudziak, and Nicholas~Donald
  Lane.
\newblock Zero-cost proxies for lightweight {NAS}.
\newblock In \emph{9th International Conference on Learning Representations,
  {ICLR} 2021, Virtual Event, Austria, May 3-7, 2021}. OpenReview.net, 2021.

\bibitem[Akhauri et~al.(2022)Akhauri, Mu{\~{n}}oz, Jain, and Iyer]{EZNAS}
Yash Akhauri, Juan~Pablo Mu{\~{n}}oz, Nilesh Jain, and Ravi Iyer.
\newblock {EZNAS:} evolving zero-cost proxies for neural architecture scoring.
\newblock In \emph{Advances in Neural Information Processing Systems 35: Annual
  Conference on Neural Information Processing Systems 2022, NeurIPS 2022, New
  Orleans, LA, USA, November 28 - December 9, 2022}, 2022.

\bibitem[Alatawi et~al.(2022)Alatawi, Alomani, Alhawiti, and
  Ayaz]{Alatawi2022PlantDD}
Anwar~Abdullah Alatawi, Shahd~Maadi Alomani, Najd~Ibrahim Alhawiti, and
  Muhammad Ayaz.
\newblock Plant disease detection using ai based vgg-16 model.
\newblock \emph{International Journal of Advanced Computer Science and
  Applications}, 2022.

\bibitem[Alowais et~al.(2023)Alowais, Alghamdi, Alsuhebany, Alqahtani, Alshaya,
  Almohareb, Aldairem, Alrashed, Bin~Saleh, Badreldin, Al~Yami, Al~Harbi, and
  Albekairy]{Alowais2023}
S.~A. Alowais, S.~S. Alghamdi, N. Alsuhebany, T. Alqahtani, A.~I. Alshaya,
  S.~N. Almohareb, A. Aldairem, M. Alrashed, K. Bin~Saleh, H.~A. Badreldin,
  M.~S. Al~Yami, S. Al~Harbi, and A.~M. Albekairy.
\newblock Revolutionizing healthcare: the role of artificial intelligence in
  clinical practice.
\newblock \emph{BMC Medical Education}, 23\penalty0 (1):\penalty0 689, 2023.

\bibitem[Camero et~al.(2018)Camero, Toutouh, and
  Alba]{CameroLowCostExpectedPerformance}
Andr{\'{e}}s Camero, Jamal Toutouh, and Enrique Alba.
\newblock Low-cost recurrent neural network expected performance evaluation.
\newblock \emph{CoRR}, abs/1805.07159, 2018.

\bibitem[Chen et~al.(2021)Chen, Gong, and Wang]{TENAS}
Wuyang Chen, Xinyu Gong, and Zhangyang Wang.
\newblock Neural architecture search on imagenet in four {GPU} hours: {A}
  theoretically inspired perspective.
\newblock In \emph{9th International Conference on Learning Representations,
  {ICLR} 2021, Virtual Event, Austria, May 3-7, 2021}. OpenReview.net, 2021.

\bibitem[Cort{\^{e}}s et~al.(2024)Cort{\^{e}}s, Louren{\c{c}}o, and
  Machado]{CortesEvoAPPS24}
Gabriel Cort{\^{e}}s, Nuno Louren{\c{c}}o, and Penousal Machado.
\newblock Towards physical plausibility in neuroevolution systems.
\newblock In \emph{Applications of Evolutionary Computation - 27th European
  Conference, EvoApplications 2024, Held as Part of EvoStar 2024, Aberystwyth,
  UK, April 3-5, 2024, Proceedings, Part {II}}, pages 76--90. Springer, 2024.

\bibitem[Dash et~al.(2022)Dash, Ansari, Sharma, and Ali]{Dash2022ThreatsAO}
Bibhu Dash, Md~Meraj Ansari, Pawankumar Sharma, and Azad Ali.
\newblock Threats and opportunities with ai-based cyber security intrusion
  detection: A review.
\newblock \emph{International Journal of Software Engineering \& Applications},
  2022.

\bibitem[{de Vries}(2023)]{DevriesEnergy}
Alex {de Vries}.
\newblock The growing energy footprint of artificial intelligence.
\newblock \emph{Joule}, 7\penalty0 (10):\penalty0 2191--2194, 2023.

\bibitem[Dong et~al.(2024)Dong, Li, Pan, Wei, Liu, Wang, and Chu]{DongParzc}
Peijie Dong, Lujun Li, Xinglin Pan, Zimian Wei, Xiang Liu, Qiang Wang, and
  Xiaowen Chu.
\newblock Parzc: Parametric zero-cost proxies for efficient {NAS}.
\newblock \emph{CoRR}, abs/2402.02105, 2024.

\bibitem[Dong and Yang(2020)]{NASBench201}
Xuanyi Dong and Yi Yang.
\newblock {NAS-Bench-201}: Extending the scope of reproducible neural
  architecture search.
\newblock In \emph{8th International Conference on Learning Representations,
  {ICLR} 2020, Addis Ababa, Ethiopia, April 26-30, 2020}. OpenReview.net, 2020.

\bibitem[Dong et~al.(2022)Dong, Liu, Musial, and Gabrys]{NATSBench}
Xuanyi Dong, Lu Liu, Katarzyna Musial, and Bogdan Gabrys.
\newblock {NATS-Bench}: Benchmarking {NAS} algorithms for architecture topology
  and size.
\newblock \emph{{IEEE} Trans. Pattern Anal. Mach. Intell.}, 44\penalty0
  (7):\penalty0 3634--3646, 2022.

\bibitem[Eiben and Smith(2015)]{Eiben2015}
A.~E. Eiben and James~E. Smith.
\newblock \emph{Introduction to Evolutionary Computing, Second Edition}.
\newblock Springer, 2015.

\bibitem[Fortin et~al.(2012)Fortin, Rainville, Gardner, Parizeau, and
  Gagn{\'{e}}]{DEAP}
F{\'{e}}lix{-}Antoine Fortin, Fran{\c{c}}ois{-}Michel~De Rainville,
  Marc{-}Andr{\'{e}} Gardner, Marc Parizeau, and Christian Gagn{\'{e}}.
\newblock {DEAP:} evolutionary algorithms made easy.
\newblock \emph{J. Mach. Learn. Res.}, 13:\penalty0 2171--2175, 2012.

\bibitem[Furman and Seamans(2018)]{AIEconomy}
Jason Furman and Robert Seamans.
\newblock Ai and the economy.
\newblock Working Paper 24689, National Bureau of Economic Research, 2018.

\bibitem[Gomes(2024)]{GomesIRobotAIEthics}
Orlando Gomes.
\newblock I, robot: the three laws of robotics and the ethics of the peopleless
  economy.
\newblock \emph{{AI} Ethics}, 4\penalty0 (2):\penalty0 257--272, 2024.

\bibitem[Google(2024)]{Google2024EnvReport}
Google.
\newblock 2024 {E}nvironmental {R}eport.
\newblock
  \url{https://sustainability.google/reports/google-2024-environmental-report/},
  2024.
\newblock [Accessed 05-10-2024].

\bibitem[Haider et~al.(2024)Haider, Zhang, Deivalaskhmi, Lakshmi~Narayanan, and
  Ko]{HaiderNeuromorphic}
Muhammad~Hamis Haider, Hao Zhang, S. Deivalaskhmi, G. Lakshmi~Narayanan, and
  Seok-Bum Ko.
\newblock \emph{Is Neuromorphic Computing the Key to Power-Efficient Neural
  Networks: A Survey}, pages 91--113.
\newblock Springer Nature Switzerland, Cham, 2024.

\bibitem[Hestness et~al.(2017)Hestness, Narang, Ardalani, Diamos, Jun,
  Kianinejad, Patwary, Yang, and
  Zhou]{hestness2017deeplearningscalingpredictable}
Joel Hestness, Sharan Narang, Newsha Ardalani, Gregory~F. Diamos, Heewoo Jun,
  Hassan Kianinejad, Md. Mostofa~Ali Patwary, Yang Yang, and Yanqi Zhou.
\newblock Deep learning scaling is predictable, empirically.
\newblock \emph{CoRR}, abs/1712.00409, 2017.

\bibitem[Kadiresan et~al.(2022)Kadiresan, Baweja, and Ogbanufe]{Kadiresan2022}
Adheesh Kadiresan, Yuvraj Baweja, and Obi Ogbanufe.
\newblock \emph{Bias in AI-Based Decision-Making}, pages 275--285.
\newblock Springer International Publishing, Cham, 2022.

\bibitem[Lee and Ham(2024)]{LeeAZNAS}
Junghyup Lee and Bumsub Ham.
\newblock {AZ-NAS:} assembling zero-cost proxies for network architecture
  search.
\newblock In \emph{{IEEE/CVF} Conference on Computer Vision and Pattern
  Recognition, {CVPR} 2024, Seattle, WA, USA, June 16-22, 2024}, pages
  5893--5903. {IEEE}, 2024.

\bibitem[Li et~al.(2023)Li, Yang, Bhardwaj, and Marculescu]{Zico}
Guihong Li, Yuedong Yang, Kartikeya Bhardwaj, and Radu Marculescu.
\newblock Zico: Zero-shot {NAS} via inverse coefficient of variation on
  gradients.
\newblock In \emph{The Eleventh International Conference on Learning
  Representations, {ICLR} 2023, Kigali, Rwanda, May 1-5, 2023}. OpenReview.net,
  2023.

\bibitem[Lin et~al.(2021)Lin, Wang, Sun, Chen, Sun, Qian, Li, and Jin]{ZenNAS}
Ming Lin, Pichao Wang, Zhenhong Sun, Hesen Chen, Xiuyu Sun, Qi Qian, Hao Li,
  and Rong Jin.
\newblock Zen-nas: {A} zero-shot {NAS} for high-performance image recognition.
\newblock In \emph{2021 {IEEE/CVF} International Conference on Computer Vision,
  {ICCV} 2021, Montreal, QC, Canada, October 10-17, 2021}, pages 337--346.
  {IEEE}, 2021.

\bibitem[Louren{\c{c}}o et~al.(2015)Louren{\c{c}}o, Pereira, and
  Costa]{LourencoSGE}
Nuno Louren{\c{c}}o, Francisco~B. Pereira, and Ernesto Costa.
\newblock {SGE:} {A} structured representation for grammatical evolution.
\newblock In \emph{Artificial Evolution - 12th International Conference,
  Evolution Artificielle, {EA} 2015, Lyon, France, October 26-28, 2015. Revised
  Selected Papers}, pages 136--148. Springer, 2015.

\bibitem[Louren{\c{c}}o et~al.(2018)Louren{\c{c}}o, Assun{\c{c}}{\~{a}}o,
  Pereira, Costa, and Machado]{LourencoDSGE}
Nuno Louren{\c{c}}o, Filipe Assun{\c{c}}{\~{a}}o, Francisco~B. Pereira, Ernesto
  Costa, and Penousal Machado.
\newblock Structured grammatical evolution: {A} dynamic approach.
\newblock In \emph{Handbook of Grammatical Evolution}, pages 137--161.
  Springer, 2018.

\bibitem[Maass(1997)]{SpikingNeurons}
Wolfgang Maass.
\newblock Networks of spiking neurons: The third generation of neural network
  models.
\newblock \emph{Neural Networks}, 10\penalty0 (9):\penalty0 1659--1671, 1997.

\bibitem[Mellor et~al.(2021)Mellor, Turner, Storkey, and Crowley]{MellorNASWOT}
Joe Mellor, Jack Turner, Amos Storkey, and Elliot~J Crowley.
\newblock Neural architecture search without training.
\newblock In \emph{Proceedings of the 38th International Conference on Machine
  Learning}, pages 7588--7598. PMLR, 2021.

\bibitem[Patterson et~al.(2022)Patterson, Gonzalez, H{\"{o}}lzle, Le, Liang,
  Munguia, Rothchild, So, Texier, and Dean]{patterson2022carbon}
David~A. Patterson, Joseph Gonzalez, Urs H{\"{o}}lzle, Quoc~V. Le, Chen Liang,
  Lluis{-}Miquel Munguia, Daniel Rothchild, David~R. So, Maud Texier, and Jeff
  Dean.
\newblock The carbon footprint of machine learning training will plateau, then
  shrink.
\newblock \emph{Computer}, 55\penalty0 (7):\penalty0 18--28, 2022.

\bibitem[Ramirez et~al.(2022)Ramirez, Kim, Hamadi, Damiani, Byon, Kim, Cho, and
  Yeun]{RamirezAttacks}
Miguel~A. Ramirez, Song{-}Kyoo Kim, Hussam M. N.~Al Hamadi, Ernesto Damiani,
  Young{-}Ji Byon, Tae{-}Yeon Kim, Chung{-}Suk Cho, and Chan~Yeob Yeun.
\newblock Poisoning attacks and defenses on artificial intelligence: {A}
  survey.
\newblock \emph{CoRR}, abs/2202.10276, 2022.

\bibitem[Robinson(2024)]{MSNMicrosoftCarbonEmissions}
Dan Robinson.
\newblock Microsoft's carbon emissions up nearly 30
\newblock
  \url{https://www.msn.com/en-us/money/other/microsofts-carbon-emissions-up-nearly-30-thanks-to-ai/ar-BB1mvgao},
  2024.
\newblock [Accessed 05-10-2024].

\bibitem[Rokh et~al.(2023)Rokh, Azarpeyvand, and
  Khanteymoori]{RokhQuantization}
Babak Rokh, Ali Azarpeyvand, and Alireza Khanteymoori.
\newblock A comprehensive survey on model quantization for deep neural networks
  in image classification.
\newblock \emph{ACM Trans. Intell. Syst. Technol.}, 14\penalty0 (6), 2023.

\bibitem[Ryan et~al.(1998)Ryan, Collins, and Neill]{RyanGE}
Conor Ryan, JJ Collins, and Michael~O. Neill.
\newblock Grammatical evolution: Evolving programs for an arbitrary language.
\newblock In \emph{Genetic Programming}, pages 83--96, Berlin, Heidelberg,
  1998. Springer Berlin Heidelberg.

\bibitem[Spector(2012)]{SpectorLexicase}
Lee Spector.
\newblock Assessment of problem modality by differential performance of
  lexicase selection in genetic programming: a preliminary report.
\newblock In \emph{Genetic and Evolutionary Computation Conference, {GECCO}
  '12, Philadelphia, PA, USA, July 7-11, 2012, Companion Material Proceedings},
  pages 401--408. {ACM}, 2012.

\bibitem[Su(2018)]{UnemploymentAI}
Grace Su.
\newblock Unemployment in the {AI} age.
\newblock \emph{AI Matters}, 3\penalty0 (4):\penalty0 35–43, 2018.

\bibitem[Tanaka et~al.(2020)Tanaka, Kunin, Yamins, and Ganguli]{Synflow}
Hidenori Tanaka, Daniel Kunin, Daniel L.~K. Yamins, and Surya Ganguli.
\newblock Pruning neural networks without any data by iteratively conserving
  synaptic flow.
\newblock In \emph{Advances in Neural Information Processing Systems 33: Annual
  Conference on Neural Information Processing Systems 2020, NeurIPS 2020,
  December 6-12, 2020, virtual}, 2020.

\bibitem[Vaddy(2023)]{Vaddy_2023}
Rama Vaddy.
\newblock {AI} and {ML} for {T}ransportation {R}oute {O}ptimization.
\newblock \emph{International Transactions in Machine Learning}, 5\penalty0
  (5):\penalty0 1–19, 2023.

\bibitem[van Wynsberghe(2021)]{SustainableAIWynsberghe}
Aimee van Wynsberghe.
\newblock Sustainable {AI:} {AI} for sustainability and the sustainability of
  {AI}.
\newblock \emph{{AI} Ethics}, 1\penalty0 (3):\penalty0 213--218, 2021.

\bibitem[Willis et~al.(1997)Willis, Hiden, Marenbach, McKay, and
  Montague]{WilliwsGPIntroSurvey}
M.-J. Willis, H.G. Hiden, P. Marenbach, B. McKay, and G.A. Montague.
\newblock Genetic programming: an introduction and survey of applications.
\newblock In \emph{Second International Conference On Genetic Algorithms In
  Engineering Systems: Innovations And Applications}, pages 314--319, 1997.

\bibitem[Wu and Tsai(2024)]{TrainingFreeNASReview}
Meng-Ting Wu and Chun-Wei Tsai.
\newblock Training-free neural architecture search: A review.
\newblock \emph{ICT Express}, 10\penalty0 (1):\penalty0 213--231, 2024.

\bibitem[Xu et~al.(2024)Xu, Yin, Cai, Yi, Xu, Wang, Wu, Zhao, Yang, Wang,
  Zhang, Lu, Zhang, Wang, Li, Liu, Jin, and Liu]{XuLLMSurvey}
Mengwei Xu, Wangsong Yin, Dongqi Cai, Rongjie Yi, Daliang Xu, Qipeng Wang,
  Bingyang Wu, Yihao Zhao, Chen Yang, Shihe Wang, Qiyang Zhang, Zhenyan Lu, Li
  Zhang, Shangguang Wang, Yuanchun Li, Yunxin Liu, Xin Jin, and Xuanzhe Liu.
\newblock A survey of resource-efficient {LLM} and multimodal foundation
  models.
\newblock \emph{CoRR}, abs/2401.08092, 2024.

\end{thebibliography}
}
\newpage
\onecolumn
\section*{Supplementary Material}

\begin{table}[h]
\centering
\caption{List of unary operations used in GreenMachine.}
\begin{tblr}{
  row{1} = {c},
  hline{1,28} = {-}{0.08em},
  hline{2} = {-}{},
}
Operation            & Description \\
Abs                  & Compute the absolute value element-wise. \\
Add noise            & Add random noise (standard normal distribution)\\& to a tensor. \\
Determinant          & Calculate the determinant of a matrix. \\
Element-wise invert       & Invert elements of a tensor element-wise. \\
Exp                  & Apply the exponential function element-wise. \\
Frobenius norm       & Compute the Frobenius norm of a tensor. \\
Gaussian initialization   & Initialize a tensor with the random values\\& from the standard normal distribution. \\
Greater than zero    & Check if elements are greater than zero. \\
Heaviside            & Compute the Heaviside step function for each element. \\
L1 norm              & Compute the L1 norm of a tensor.\\
Less than zero       & Check if elements are less than zero. \\
Log                  & Apply the natural logarithm element-wise. \\
Log determinant              & Compute the log determinant of a matrix. \\
Normalized sum       & Return the sum of tensor elements normalized. \\
Normalization        & Scale tensor values to a 0-1 range. \\
Num elements         & Return the number of elements in a tensor. \\
Ones like            & Create a tensor of ones with the same shape as the input. \\
ReLu                 & Apply the ReLU activation element-wise. \\
Sigmoid              & Apply the sigmoid function element-wise. \\
Sign                 & Extract the sign of each element. \\
Softmax              & Apply the softmax function element-wise. \\
Squared power        & Raise each element to the power of 2. \\
Transpose            & Compute the transpose of a matrix. \\
Zeros like           & Create a tensor of zeros with the same shape as the input. \\
               &                                                                                             
\end{tblr}
\end{table}

\begin{table}[h]
\centering
\caption{List of binary operations used in GreenMachine.}
\begin{tblr}{
  row{1} = {c},
  hline{1,13} = {-}{0.08em},
  hline{2} = {-}{},
}
Operation                & Description \\
Cosine similarity        & Calculates cosine similarity between tensors. \\
Element-wise product     & Multiplies tensors element-wise. \\
Equal                    & Checks element-wise equality of tensors. \\
Greater than             & Checks element-wise if one tensor is greater than another. \\
Kullback–Leibler divergence & Computes the Kullback-Leibler divergence. \\
Less than                & Checks element-wise if one tensor is less than another. \\
Matrix multiplication    & Performs matrix multiplication. \\
Max                      & Returns the element-wise maximum of two tensors. \\
Min                      & Returns the element-wise minimum of two tensors. \\
Subtraction              & Subtracts one tensor from another element-wise. \\
Sum                      & Adds two tensors element-wise. \\
               &                                                                                             
\end{tblr}
\end{table}

\begin{figure*}[ht]
    \small
    {
    \begin{verbatim}
    GreenMachine-1: (greater_than((mat_mul(pass_noise_wt, (greater_than((
            kl_div(pass_noise_grad, pass_noise_wt)), 
            subtract(pass_noise_fwd_output, pass_perturbation_bwd_input))))), random_wt))
    
    GreenMachine-2: max(softmax(pass_perturbation_fwd_output), equal(pass_perturbation_grad, 
            element_wise_product(pass_perturbation_bwd_input, pass_bwd_input)))
    
    GreenMachine-3: cosine_similarity(softmax(cosine_similarity(pass_perturbation_bwd_output, 
            pass_fwd_output)),(greater_than(greater_than(pass_noise_fwd_output, cosine_similarity(
            pass_perturbation_grad, 
            less_than(less_than(equal((equal(abs((max(transpose(pass_noise_bwd_output), 
            normalize(less_than_zero(pass_fwd_output))))), (greater_than(gaussian_init((
            kl_div(power(pass_noise_bwd_input), 
            normalized_sum(pass_grad)))), element_wise_product(kl_div( 
            cosine_similarity(pass_perturbation_fwd_output, random_grad), 
            (sum(random_wt, pass_perturbation_grad))),  
            greater_than(pass_bwd_output, (max(pass_perturbation_fwd_output, pass_bwd_input)))))))), 
            (subtract(element_wise_invert(pass_perturbation_wt),
            gaussian_init(determinant(pass_perturbation_bwd_input))))), 
            (greater_than(kl_div((min(random_grad, pass_noise_bwd_input)), pass_noise_fwd_output), 
            (mat_mul((element_wise_product(pass_noise_fwd_input, pass_perturbation_fwd_output)), 
            ones_like(sum((mat_mul(frobenius_norm(pass_fwd_input), 
            frobenius_norm(pass_perturbation_bwd_output))), numel(l1_norm(pass_noise_grad))))))))), 
            greater_than(pass_noise_fwd_input, (subtract(sum(random_wt, 
            heaviside(pass_noise_bwd_output)), softmax((mat_mul(sum(greater_than(pass_fwd_output, 
            cosine_similarity(pass_perturbation_bwd_output, pass_perturbation_fwd_input)), 
            (sum((min(pass_bwd_output, pass_noise_wt)), 
            cosine_similarity(pass_grad, pass_noise_grad)))),
            sigmoid(frobenius_norm(normalize(pass_fwd_output)))))))))))), random_wt)))
    \end{verbatim}
    }
    \caption{Formulas of the best solutions found.}
    \label{fig:best_solutions}
\end{figure*}

\begin{table*}[ht]
\centering
\caption{Comparison of Zero-Cost proxies on the NATS-Bench benchmark across the CIFAR-10, CIFAR-100, and ImageNet16-120 datasets on the \textbf{non-stratified subset}. The presented values are the \textbf{mean absolute Spearman correlation coefficient} over 30 runs, multiplied by 100. Bold denotes the best value.}
\label{table:resultsnon_stratified_spearman}
\begin{tblr}{
  colsep=4pt,
  rows={0.5mm},
  hline{1,2,14} = {-}{0.08em},
  hline{11} = {-}{},
  colspec = {Q[l] Q[c] Q[c] Q[c] Q[c] Q[c] Q[c]},
}
Proxy & SSS (CF-10) & SSS (CF-100) & SSS (IN16-120) & TSS (CF-10) & TSS (CF-100) & TSS (IN16-120)  &        \\
 Random	& 5.9 $\pm$ 3.3	& 6.6 $\pm$ 5.7	& 6.8 $\pm$ 5.4	& 6.4 $\pm$ 5.1	& 6.2 $\pm$ 4.3	& 6.1 $\pm$ 4.7\\
\#Params	& 85.6 $\pm$ 2.8	& 71.5 $\pm$ 5.4	& 82.9 $\pm$ 3.4	& 51.9 $\pm$ 5.9	& 48.9 $\pm$ 7.3	& 39.3 $\pm$ 6.7\\
Synflow \cite{Synflow}	& \textbf{92.6 $\pm$ 1.3}	& 78.2 $\pm$ 4.1	& \textbf{93.2 $\pm$ 1.6}	& 52.8 $\pm$ 7.4	& 50.1 $\pm$ 7.6	& 50.0 $\pm$ 6.3\\
Gradnorm \cite{Gradnorm}	& 27.6 $\pm$ 7.5	& 67.0 $\pm$ 5.9	& 52.4 $\pm$ 7.5	& 21.1 $\pm$ 8.4	& 8.7 $\pm$ 6.2	& 9.5 $\pm$ 5.4\\
NASWOT \cite{MellorNASWOT}	& 54.0 $\pm$ 5.3	& 23.5 $\pm$ 7.6	& 56.6 $\pm$ 5.5	& 60.0 $\pm$ 5.7	& 54.0 $\pm$ 5.6	& 54.2 $\pm$ 7.4\\
TE-NAS \cite{TENAS}	& 48.9 $\pm$ 5.5	& 40.1 $\pm$ 8.0	& 54.3 $\pm$ 7.3	& 32.8 $\pm$ 7.1	& 16.6 $\pm$ 8.1	& 11.6 $\pm$ 8.0\\
Zen-NAS \cite{ZenNAS}	& 89.7 $\pm$ 1.8	& 64.9 $\pm$ 6.1	& 84.2 $\pm$ 2.8	& 17.8 $\pm$ 7.6	& 24.4 $\pm$ 8.1	& 31.1 $\pm$ 6.9\\
ZiCo \cite{Zico}	& 90.2 $\pm$ 1.7	& 75.7 $\pm$ 3.3	& 89.9 $\pm$ 2.1	& 54.9 $\pm$ 6.9	& 51.2 $\pm$ 7.0	& 51.1 $\pm$ 6.3\\
EZNAS \cite{EZNAS}	& 89.6 $\pm$ 1.9	& 67.0 $\pm$ 4.9	& 76.0 $\pm$ 3.0	& \textbf{79.6 $\pm$ 3.3}	& \textbf{78.7 $\pm$ 3.8}	& \textbf{72.4 $\pm$ 4.8}\\
GM-1 (Ours)	& 70.4 $\pm$ 3.5	& 72.2 $\pm$ 4.9	& 68.6 $\pm$ 4.4	& 67.6 $\pm$ 5.6	& 65.1 $\pm$ 5.4	& 54.4 $\pm$ 5.9\\
GM-2 (Ours)	& 86.2 $\pm$ 2.6	& 80.4 $\pm$ 3.7	& 93.1 $\pm$ 1.0	& 47.3 $\pm$ 7.0	& 44.4 $\pm$ 7.0	& 44.8 $\pm$ 6.3\\
GM-3 (Ours)	& 91.2 $\pm$ 1.9	& \textbf{84.2 $\pm$ 3.2}	& 87.7 $\pm$ 2.3	& 13.7 $\pm$ 7.0	& 10.6 $\pm$ 6.2	& 40.4 $\pm$ 6.9
\end{tblr}
\end{table*}

\begin{table*}[ht]
\centering
\caption{Comparison of Zero-Cost proxies on the NATS-Bench benchmark across the CIFAR-10, CIFAR-100, and ImageNet16-120 datasets on the \textbf{stratified subset}. The presented values are the \textbf{mean absolute Spearman correlation coefficient} over 30 runs, multiplied by 100. Bold denotes the best value.}
\label{table:resultsstratified_spearman}
\begin{tblr}{
  colsep=4pt,
  rows={0.5mm},
  hline{1,2,14} = {-}{0.08em},
  hline{11} = {-}{},
  colspec = {Q[l] Q[c] Q[c] Q[c] Q[c] Q[c] Q[c]},
}
Proxy & SSS (CF-10) & SSS (CF-100) & SSS (IN16-120) & TSS (CF-10) & TSS (CF-100) & TSS (IN16-120)  &        \\
 Random	& 7.5 $\pm$ 5.0	& 6.6 $\pm$ 4.4	& 6.2 $\pm$ 5.3	& 6.4 $\pm$ 4.8	& 5.0 $\pm$ 4.3	& 6.0 $\pm$ 4.4\\
\#Params	& 91.6 $\pm$ 1.7	& 66.4 $\pm$ 5.4	& 76.9 $\pm$ 3.8	& 51.1 $\pm$ 4.9	& 58.4 $\pm$ 6.0	& 43.6 $\pm$ 5.5\\
Synflow \cite{Synflow}	& 94.3 $\pm$ 1.1	& 78.4 $\pm$ 4.0	& 87.0 $\pm$ 2.3	& 91.1 $\pm$ 1.2	& 65.2 $\pm$ 3.8	& 55.8 $\pm$ 5.7\\
Gradnorm \cite{Gradnorm}	& 59.5 $\pm$ 5.0	& 80.9 $\pm$ 1.7	& 67.1 $\pm$ 4.4	& 79.0 $\pm$ 2.4	& 30.3 $\pm$ 6.9	& 10.5 $\pm$ 7.4\\
NASWOT \cite{MellorNASWOT}	& 73.0 $\pm$ 3.9	& 21.9 $\pm$ 10.3	& 48.8 $\pm$ 6.8	& 88.7 $\pm$ 2.0	& 72.1 $\pm$ 3.6	& 60.0 $\pm$ 4.8\\
TE-NAS \cite{TENAS}	& 74.3 $\pm$ 3.2	& 40.9 $\pm$ 8.6	& 51.0 $\pm$ 6.4	& 31.3 $\pm$ 6.6	& 26.5 $\pm$ 10.1	& 15.8 $\pm$ 7.3\\
Zen-NAS \cite{ZenNAS}	& 92.5 $\pm$ 1.2	& 64.7 $\pm$ 5.6	& 76.4 $\pm$ 3.8	& 48.5 $\pm$ 6.8	& 27.0 $\pm$ 8.9	& 32.2 $\pm$ 8.0\\
ZiCo \cite{Zico}	& 93.9 $\pm$ 1.1	& 76.8 $\pm$ 3.8	& 94.9 $\pm$ 0.5	& 91.0 $\pm$ 1.4	& 67.3 $\pm$ 4.1	& 80.4 $\pm$ 2.7\\
EZNAS \cite{EZNAS}	& 95.9 $\pm$ 0.4	& 82.9 $\pm$ 2.3	& 88.8 $\pm$ 1.6	& 76.1 $\pm$ 2.4	& \textbf{88.5 $\pm$ 1.5}	& \textbf{84.7 $\pm$ 3.1}\\
GM-1 (Ours)	& 81.6 $\pm$ 2.5	& 84.7 $\pm$ 2.3	& 86.8 $\pm$ 1.6	& \textbf{92.8 $\pm$ 1.7}	& 87.3 $\pm$ 2.0	& 82.7 $\pm$ 2.8\\
GM-2 (Ours)	& 95.8 $\pm$ 0.7	& \textbf{92.4 $\pm$ 0.9}	& \textbf{97.1 $\pm$ 0.3}	& 83.1 $\pm$ 2.8	& 74.1 $\pm$ 3.5	& 73.3 $\pm$ 4.0\\
GM-3 (Ours)	& \textbf{98.2 $\pm$ 0.3}	& 90.0 $\pm$ 1.0	& 94.2 $\pm$ 0.7	& 57.8 $\pm$ 5.5	& 44.8 $\pm$ 6.0	& 74.8 $\pm$ 3.6
\end{tblr}
\end{table*}

\end{document}